\documentclass[10pt,twocolumn,letterpaper]{article}

\usepackage{iccv}
\usepackage{times}
\usepackage{epsfig}
\usepackage{graphicx}
\usepackage{amsmath}
\usepackage{amssymb}

\usepackage{subfigure}
\usepackage{bm}
\usepackage{amsmath}
\usepackage[utf8]{inputenc}
\usepackage[english]{babel}
\usepackage{comment}
\usepackage{color}
\usepackage{etoolbox}
\usepackage{booktabs}
\usepackage{multirow}
\usepackage{dblfloatfix}

\newbool{inccomment}
\setbool{inccomment}{true}
\newtheorem{theorem}{Theorem}
\newcommand{\hl}[1]{\ifbool{inccomment}{{\color{magenta}#1}}{}}
\newcommand{\ww}[1]{\ifbool{inccomment}{{\color{blue} #1}}{}}
\newcommand{\yc}[1]{\ifbool{inccomment}{{\color{red} #1}}{}}

\usepackage[pagebackref=true,breaklinks=true,letterpaper=true,colorlinks,bookmarks=false]{hyperref}

\iccvfinalcopy 

\def\iccvPaperID{279} 

\ificcvfinal\fi
\begin{document}

\title{Coordinating Filters for Faster Deep Neural Networks}

\author{Wei Wen\\
	University of Pittsburgh\\
	{\tt\small wew57@pitt.edu}
	\and
	Cong Xu\\
	Hewlett Packard Labs\\
	{\tt\small cong.xu@hpe.com}
	\and
	Chunpeng  Wu\\
	University of Pittsburgh\\
	{\tt\small chw127@pitt.edu}
	\and
	Yandan  Wang\\
	University of Pittsburgh\\
	{\tt\small yaw46@pitt.edu}
	\and
	Yiran  Chen\\
	Duke University\\
	{\tt\small yiran.chen@duke.edu}
	\and
	Hai  Li\\
	Duke University\\
	{\tt\small hai.li@duke.edu}
}

\maketitle
\thispagestyle{empty}

\begin{abstract}
   Very large-scale \textit{Deep Neural Networks} (DNNs) have achieved remarkable successes in a large variety of computer vision tasks. 
   However, the high computation intensity of DNNs makes it challenging to deploy these models on resource-limited systems.
   Some studies used low-rank approaches that approximate the filters by low-rank basis to accelerate the testing.
   Those works directly decomposed the pre-trained DNNs by Low-Rank Approximations (LRA). 
   How to train DNNs toward lower-rank space for more efficient DNNs, however, remains as an open area.
   To solve the issue, in this work, we propose Force Regularization, which uses attractive forces to enforce filters so as to coordinate more weight information into lower-rank space\footnote{The source code is available in \url{https://github.com/wenwei202/caffe}}. 
   We mathematically and empirically verify that after applying our technique, standard LRA methods can reconstruct filters using much lower basis and thus result in faster DNNs. 
   The effectiveness of our approach is comprehensively evaluated in \textit{ResNets}, \textit{AlexNet}, and \textit{GoogLeNet}. 
   In \textit{AlexNet}, for example, Force Regularization gains $2\times$ speedup on modern GPU without accuracy loss and $4.05\times$ speedup on CPU by paying small accuracy degradation.
   Moreover, Force Regularization better initializes the low-rank DNNs such that the fine-tuning can converge faster toward higher accuracy. 
   The obtained lower-rank DNNs can be further sparsified, proving that \textit{Force Regularization} can be integrated with state-of-the-art sparsity-based acceleration methods.
\end{abstract}

\section{Introduction}
\label{sec:intro}

\textit{Deep Neural Networks} (DNNs) have achieved record-breaking accuracy in many image classification tasks~\cite{Alex_NIPS2012_4824}
\cite{Vggnet_2014}\cite{GoogleNet_2015}\cite{Kaiming_ResNet_ICCV}. 
With the advances of algorithms, availability of database, and improvement in hardware performance, the depth of DNNs grows dramatically from a few to hundreds or even thousands of layers, enabling human-level performance~\cite{he2015delving}. 
However, deploying these large models on resource-limited platforms, \textit{e.g.,} mobiles and autonomous cars, is very challenging due to the high demand in the computation resource and hence energy consumption.

Recently, many techniques to accelerate the testing process of deployed DNNs have been studied, such as 
weight sparsifying or connection pruning~\cite{Han_NIPS2015}\cite{Han_ICLR2016}\cite{Wen_NIPS2016}\cite{Jongsoo_ICLR2017}\cite{Liu_CVPR2015}\cite{guo_dnf_nips2016} \cite{Lebedev_2016_CVPR}.
These approaches require delicate hardware customization and/or software design to transfer sparsity into practical speedup. 
Unlike sparsity-based methods, \textit{Low-Rank Approximation} (LRA) methods \cite{Liu_CVPR2015}\cite{Denil_NIPS2013_5025}\cite{Denton_NIPS2014}\cite{Max_J_arxiv2014}\cite{Yani_arxiv_2015} \cite{Cheng_arxiv_2015}\cite{wang2016accelerating}\cite{lebedev2014speeding}\cite{7332968}\cite{kim2015compression} directly decompose an original large model to a compact model with more lightweight layers.
Thanks to the redundancy (correlation) among filters in DNNs, original weight tensors can be approximated by very low-rank basis.
From the viewpoint of matrix computation, LRA approximates a large weight matrix by the product of two or more small ones to reduce computation complexity.

\begin{figure}[t]
	\centering
	
	\subfigure
	{
		\includegraphics[width=1.0\columnwidth]{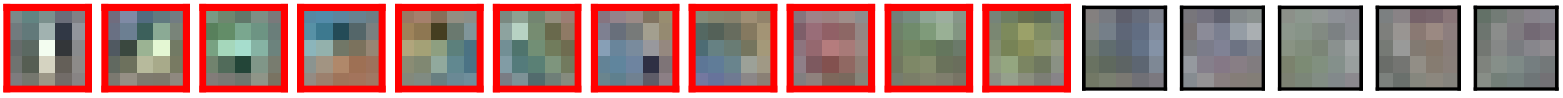}
	}
	\vfill
	\vspace{-10pt}
	\subfigure
	{
		\includegraphics[width=1.0\columnwidth]{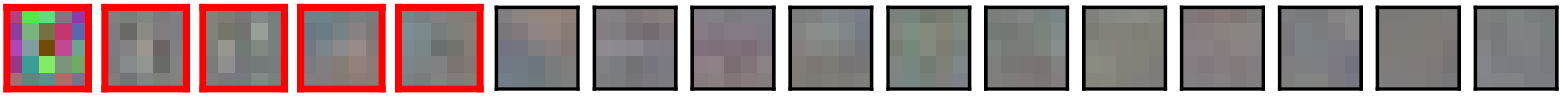}
	}
	\caption{The low-rank basis of filters in the first layer of the convolutional neural network \cite{Alex_NIPS2012_4824} on CIFAR-10. The low-rank basis is formed by the most significant principal filters that are obtained by PCA. 
	Top: the low-rank basis of the original network. 
	Bottom: the low-rank basis of the same network after applying \textit{Force Regularization}. 
	The number of red boxes indicates the required rank to reconstruct the original filters with $\leq20\%$ error.}
	
	\label{fig:cifar10_conv1_filters}
\end{figure}

Previous LRA methods mostly focus on how to decompose the pre-trained weight tensors for maximizing the reduction of computation complexity, meanwhile retaining the classification accuracy. 
Instead, we propose to nudge the weights by additional gradients (\textit{attractive forces}) to coordinate the filters to a more correlated state.
Our approach aims to improve the correlation among filters and therefore obtain more lightweight DNNs through LRA.
\textit{To the best of our knowledge, this is the first work to train DNNs toward lower-rank space such that LRA can achieve faster DNNs.}


The motivation of this work is fundamental. 
It has been proven that trained filters are highly clustered and correlated \cite{Denton_NIPS2014}\cite{Denil_NIPS2013_5025}\cite{Max_J_arxiv2014}. 
Suppose each filter is reshaped as a vector.
A cluster of highly-correlated vectors then will have small included angles. 
If we are able to coordinate these vectors toward a state with smaller included angles, the correlation of the filters within that cluster improves. 
Consequently, LRA can produce a DNN with lower ranks and higher computation efficiency.

We propose a \textit{Force Regularization} to coordinate filters in DNNs. 
As demonstrated in Fig.~\ref{fig:cifar10_conv1_filters}, when using the same LRA method, say, cross-filter \textit{Principal Component Analysis} (PCA)~\cite{7332968}, applying \textit{Force Regularization} can greatly reduce the required ranks from the original design (\textit{i.e.}, $5$~\textit{vs.}~$11$),
while keeping the same approximation errors ($\leq 20\%$).
As we shall show in Section~\ref{sec:exp}, applying \textit{Force Regularization} in the training of state-of-the-art DNNs will successfully obtain lower-rank DNNs and thus improve computation efficiency, \textit{e.g.}, $4.05\times$ speedup for \textit{AlexNet} with small accuracy loss.

The contributions of our work include:
(1) We propose an effective and easy-to-implement \textit{Force Regularization} to train DNNs for \textit{lower}-rank approximation. To the best of our knowledge, this is the first work to manipulate the correlation among filters during training such that LRA can achieve faster DNNs; 
(2) DNNs manipulated by \textit{Force Regularization} can have better initialization for the retraining of LRA-decomposed DNNs, resulting in faster convergence to better accuracy;
(3) Those lightweight DNNs that have been aggressively compressed by our method can be further sparsified. 
That is, our method can be integrated with state-of-the-art sparsity-based methods to potentially achieve faster computation;
(4) \textit{Force Regularization} can be easily generalized to \textit{Discrimination Regularization} that can learn more discriminative filters to improve classification accuracy;
(5) Our implementation is open-source on both CPUs and GPUs.
\section{Related work}
\label{sec:related}


\textbf{Low-rank approximation.} 
LRA method decomposes a large model to a compact one with more lightweight layers by weight/tensor factorization. Denil~\textit{et al.}~\cite{Denil_NIPS2013_5025} studied different dictionaries to remove the redundancy between filters and channels in DNNs. 
Jaderberg~\textit{et al.}~\cite{Max_J_arxiv2014} explored filter and data reconstruction optimizations to attain optimal separable basis. 
Denton~\textit{et al.}~\cite{Denton_NIPS2014} clustered filters, extended LRA (\textit{e.g.}, \textit{Singular Value Decomposition}, SVD ) to larger-scale DNNs, and achieved $2\times$ speedup for the first two layers with $1\%$ accuracy loss.
Many new decomposition methods were proposed~\cite{Yani_arxiv_2015}\cite{Cheng_arxiv_2015}\cite{lebedev2014speeding}\cite{7332968} and the effectiveness of LRA in state-of-the-art DNNs were evaluated~\cite{Vggnet_2014}\cite{GoogleNet_2015}. 
Similar evaluations on mobile devices were also reported~\cite{kim2015compression}\cite{wang2016accelerating}.
Unlike them, we propose \textit{Force Regularization} to coordinate DNN filters to more correlated states, in which \textit{lower}-rank or more compact DNNs are achievable for faster computation.

\textbf{Sparse deep neural networks.}  The studies on sparse DNNs can be categorized into two types: non-structured~\cite{lecun1989optimal}\cite{Jongsoo_ICLR2017}\cite{Liu_CVPR2015}\cite{Han_NIPS2015}\cite{guo_dnf_nips2016} and structured~\cite{Wen_NIPS2016}\cite{Haoli_ICLR2017}\cite{Lebedev_2016_CVPR}\cite{alvarez2016learning} sparsity methods.
The first category prunes each connection independently. Consequently, sparse weights are randomly distributed. The level of non-structured sparsity is usually insufficient to achieve good practical speedup in modern hardware~\cite{Wen_NIPS2016}\cite{Lebedev_2016_CVPR}. 
Software optimization~\cite{Jongsoo_ICLR2017}\cite{Liu_CVPR2015} and hardware customization~\cite{Han_ICLR2016} are proposed to overcome this issue.
Conversely, the structured approaches prune connections group by group, such that the sparsified DNNs have regular distribution of sparse weights. The regularity is friendly to modern hardware for acceleration.
Our work is orthogonal to sparsity-based methods. 
More importantly, we find that DNNs accelerated by our method can be further sparsified by both non-structured and structured sparsity methods, potentially achieving faster computation.
\section{Correlated Filters and Their Approximation}
\label{sec:lra}

\begin{figure}[b]
	\centering
	
	\subfigure
	{
		\includegraphics[width=.45\columnwidth]{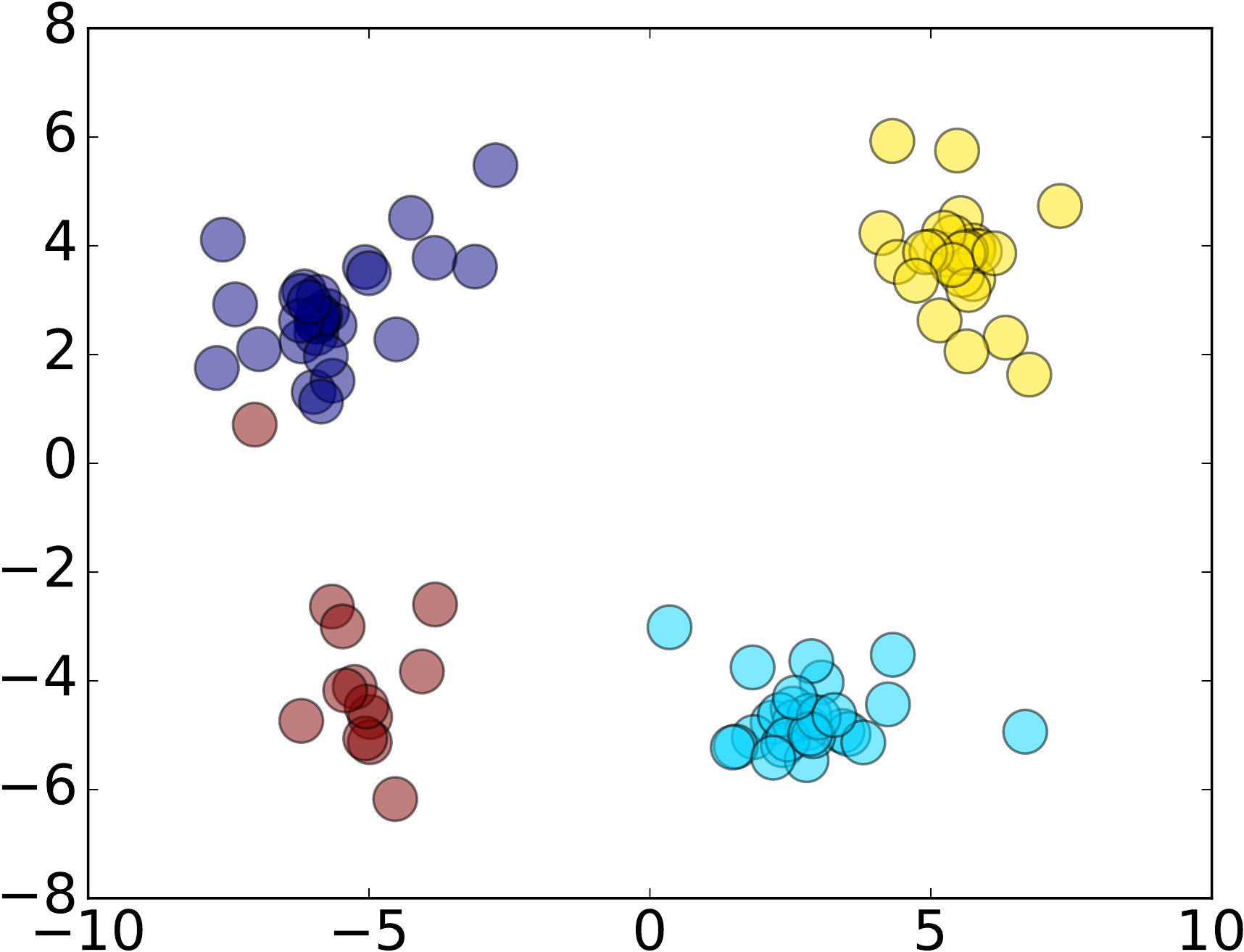}
	}
	\subfigure
	{
		\includegraphics[width=.45\columnwidth]{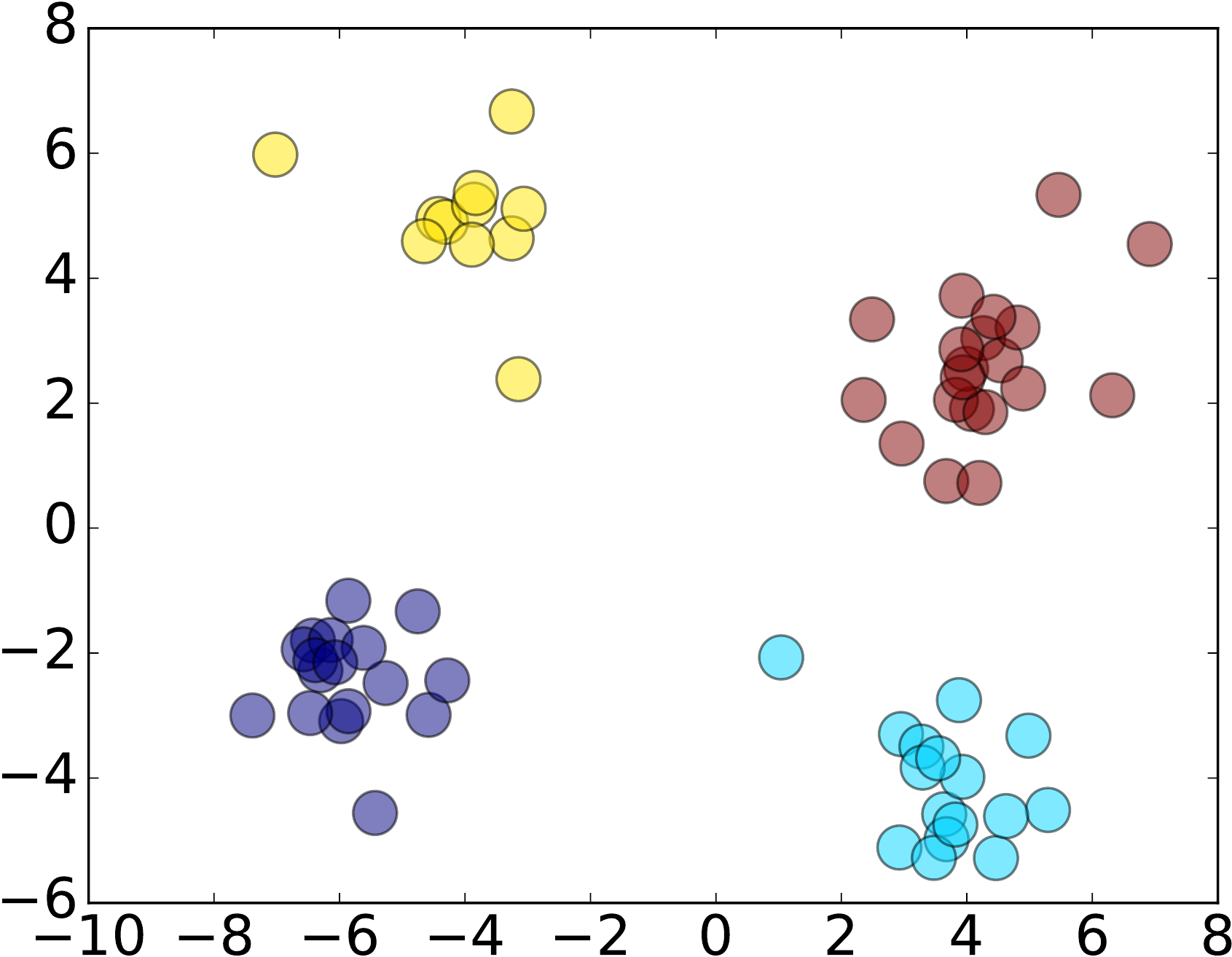}
	}
	\caption{\textit{Linear Discriminant Analysis} (LDA) of filters in the first convolutional layer of \textit{AlexNet} (left) and \textit{GoogLeNet} (right).}
	\label{fig:conv1_lda}
\end{figure}

The prior knowledge is that correlation exists among trained filters in DNNs and those filters lie in a low-rank space. For example, the color-agnostic filters \cite{Alex_NIPS2012_4824} learned in the first layer of \textit{AlexNet} lie in a hyper-plane, where RGB channels at each pixel have the same value.  Fig.~\ref{fig:conv1_lda} presents the results of \textit{Linear Discriminant Analysis} (LDA) of the first convolutional filters in \textit{AlexNet} and \textit{GoogLeNet}. The filters are normalized to unit vectors and colored to four clusters by k-means clustering, and then projected to 2D space by LDA to maximize cluster separation. The figure indicates high correlation among filters within a cluster. 
A na\"{\i}ve approach of filter approximation is to use the centroid of a cluster to approximate filters within that cluster, thus, the number of clusters is the rank of the space.
Essentially, k-means clustering is a LRA~\cite{bauckhage2015kmeans} method, although we will later show that other LRA methods can give better approximation.
The motivation of this work is that if we are able to nudge filters during the training such that the filters within a cluster are coordinated closer and some adjacent clusters are even merged into one cluster, then more accurate filter approximation using lower rank can be achieved.
We propose \textit{Force Regularization} to realize it.

Before introducing \textit{Force Regularization}, we first mathematically formulate LRA of DNN filters.
Theoretically, almost all LRA methods can gain lower-rank approximation upon our method because filters are coordinated to more correlated state.
Instead of onerously replicating all of these LRA methods, we choose cross-filter approximation~\cite{Denil_NIPS2013_5025}\cite{7332968} and a state-of-the-art work in~\cite{Cheng_arxiv_2015} as our baselines.

\begin{figure}
	\centering
	\includegraphics[width=1.0\columnwidth]{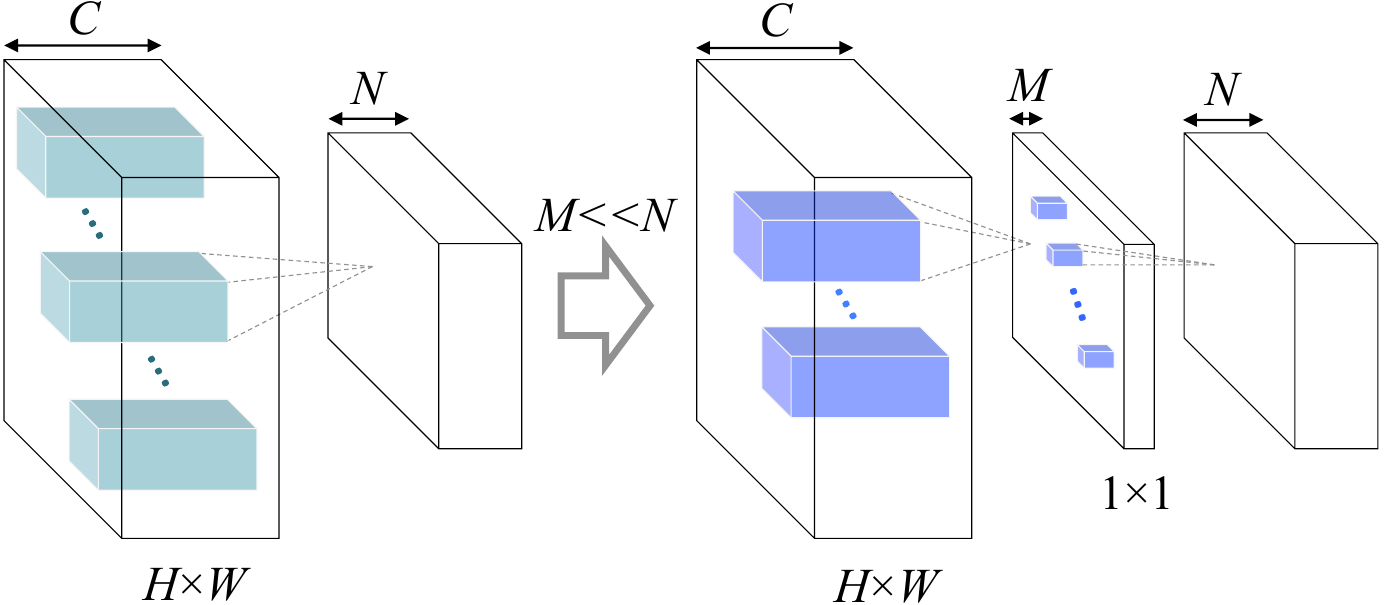}
	\caption{Cross-filter LRA of a convolutional layer.}
	\label{fig:lra_3d_filters}	
\end{figure} 

Fig.~\ref{fig:lra_3d_filters} illustrates the cross-filter approximation of a convolutional layer.
We assume all weights in a convolutional layer is a tensor $\mathcal{W}\in\mathbb{R}^{N \times C \times H \times W}$,
where $N$ and $C$ are the numbers of filters and input channels, and $H$ and $W$ are the spatial height and width of the filters, respectively.
With input feature map $\mathcal{I}$, the $n$-th output feature map $\mathcal{O}_n=\mathcal{W}_n \ast \mathcal{I}$, where $\mathcal{W}_n\in\mathbb{R}^{1 \times C \times H \times W}$ is the $n$-th filter.
Because of the redundancy (or correlation) across the filters~\cite{Denil_NIPS2013_5025}, tensor $\mathcal{W}_n (\forall n\in [1...N])$ can be approximated by a linear combination of the basis $\mathcal{B}_m\in\mathbb{R}^{1 \times C \times H \times W} (m\in [1...M], M \ll N)$ of a low-rank space $\mathcal{B}\in\mathbb{R}^{M \times C \times H \times W}$, such as
\begin{equation}
\mathcal{O}_n
\approx \left(\sum_{m=1}^{M}b^{(n)}_m\mathcal{B}_m\right)  \ast \mathcal{I}
= \sum_{m=1}^{M}\left( b^{(n)}_m \mathcal{F}_m \right).
\label{eq:cross_filter_lra}
\end{equation}
Where $b^{(n)}_m$ is a scalar, and $\mathcal{F}_m = \mathcal{B}_m \ast \mathcal{I}$ is the feature map generated by basis filter $\mathcal{B}_m$. 
Therefore, the output feature map $\mathcal{O}_n$ is a linear combination of $\mathcal{F}_m (m \in [1...M])$ which can be interpreted as the feature map basis.
Since the linear combination essentially is a $1\times1$ convolution, the convolutional layer can be decomposed to two sequential lightweight convolutional layers as shown in Fig.~\ref{fig:lra_3d_filters}. The original computation complexity is $\mathcal{O}(NCHWH^{'}W^{'})$, where $H^{'}$ and $W^{'}$ is the height and width of output feature maps, respectively. After applying cross-filter LRA, the computation complexity is reduced to $\mathcal{O}(MCHWH^{'}W^{'}+NMH^{'}W^{'})$. The computation complexity decreases when the rank $M < \frac{NCHW}{CHW+N}$.


\section{Force Regularization}
\label{sec:method}

\subsection{Regularization by Attractive Forces}
\label{method:similarity_reg}
This section proposes \textit{Force Regularization} from the perspective of physics. 
It is a gradient-based approach that adds extra gradients to data loss gradients. 
The data loss gradients aim to minimize classification error as traditional DNNs do. 
The extra gradients introduced by \textit{Force Regularization} gently adjust the lengths and directions of data loss gradients so as to nudge filters to a more correlated state. 
With a good setup of hyper-parameter, our method can coordinate more useful information of filters to a lower-rank space meanwhile maintain accuracy.
Inspired by Newton's Laws, we propose an intuitive, computation-efficient and effective \textit{Force Regularization} that uses attractive forces to coordinate filters.

\begin{figure}
	\centering
	\includegraphics[width=.6\columnwidth]{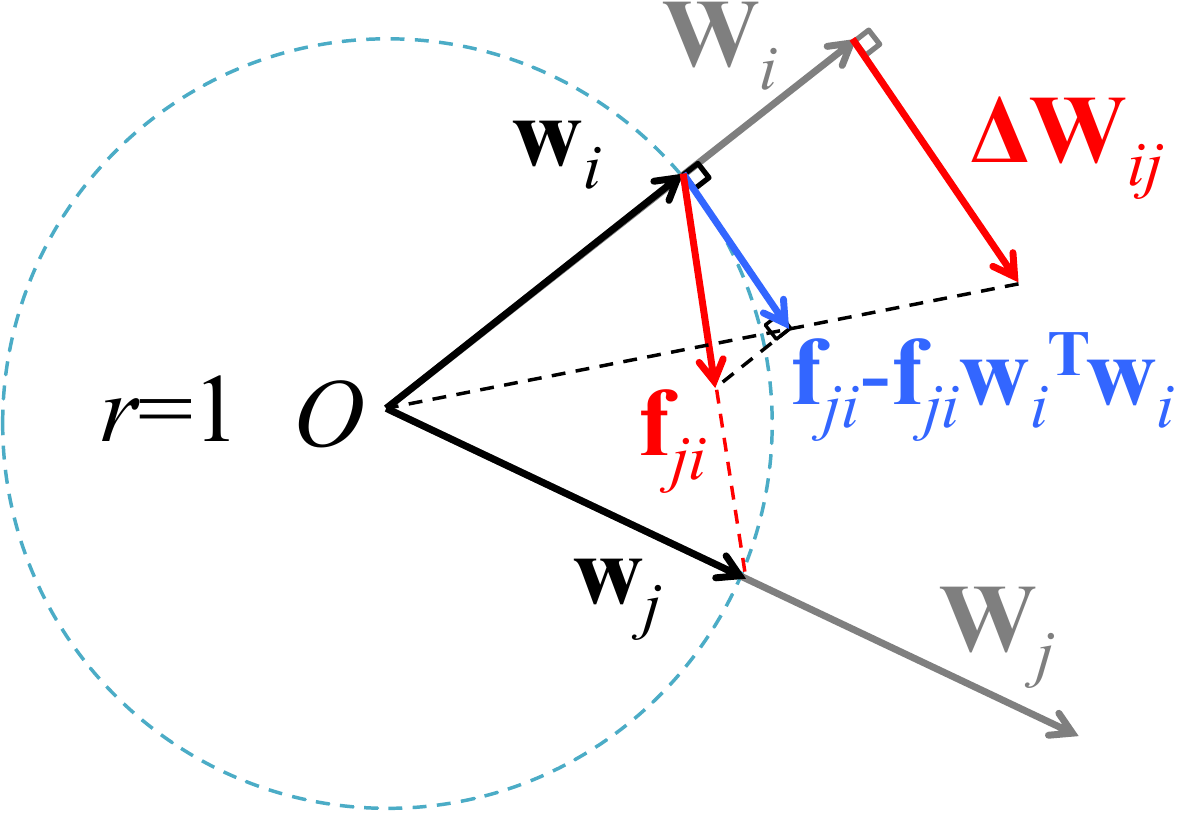}
	\caption{\textit{Force Regularization} to coordinate filters.}
	\label{fig:similarity_reg}
\end{figure}

\textit{\textbf{Force Regularization}}: 
As illustrated in Fig.~\ref{fig:similarity_reg}, suppose the filter $\mathcal{W}_n \in \mathcal{W}$ is reshaped as a vector $\mathbf{W}_{n}\in\mathbb{R}^{1 \times CHW}$ and normalized as $\mathbf{w}_{n}~\in~\mathbb{R}^{1 \times CHW}~(\forall n\in [1...N]) $, with their origin at $O$. 
We introduce the pair-wise \textit{attractive force} $\mathbf{f}_{ji}=f(\mathbf{w}_j - \mathbf{w}_i)~(\forall i,j\in [1...N])$ on $\mathbf{w}_i$ generated by $\mathbf{w}_j$. 
The gradient of \textit{Force Regularization} to update filter $\mathbf{W}_i$ is defined as
\begin{equation}
\begin{split}
\Delta\mathbf{W}_{i} 
& =  \sum_{j=1}^{N} \Delta\mathbf{W}_{ij} = ||\mathbf{W}_{i}||  \sum_{j=1}^{N} \left(\mathbf{f}_{ji} - \mathbf{f}_{ji} \mathbf{w}^T_i  \mathbf{w}_{i} \right) ,
\label{eq:simi_reg}
\end{split}
\end{equation}
where $||~\cdot~||$ is the Euclidean norm. 
The regularization gradient in Eq.~(\ref{eq:simi_reg}) is perpendicular to filter vector and can be efficiently computed by addition and multiplication.
The final updating of weights by gradient descent is
\begin{equation}
\mathbf{W}_{i} \leftarrow \mathbf{W}_{i} - \eta \cdot \left( \frac{\partial E(\mathcal{W})}{\partial \mathbf{W}_i} - \lambda_s \cdot \Delta \mathbf{W}_{i} \right),
\label{eq:sgd}
\end{equation}
where $E(\mathcal{W})$ is data loss, $\eta$ is learning rate and $\lambda_s>0$ is the coefficient of \textit{Force Regularization} to trade off the rank and accuracy. We select $\lambda_s$ by cross-validation in this work. The gradient of common weight-wise regularization (\textit{e.g.,} $\ell_2$-norm) is omitted in Eq.~(\ref{eq:sgd}) for simplicity.

Fig.~\ref{fig:similarity_reg} intuitively explains our method. Suppose each vector $\mathbf{w}_i$ is a rigid stick and there is a particle fixed at the endpoint. The particle has unit mass, and the stick is massless and can freely spin around the origin. Given the pair-wise attractive forces (\textit{e.g.,} universal gravitation) $\mathbf{f}_{ji}$, Eq.~(\ref{eq:simi_reg}) is the acceleration of particle $i$. As the forces are attractive, neighbor particles tend to spin around the origin to assemble together.
Although our regularizer seems to collapse all particles to one point which is the rank-one space for most lightweight DNNs, there exist gradients of data loss to avoid this. 
More specific, pre-trained filters orient to discriminative directions $\textbf{w}_n~(n \in [1...N])$. 
In each direction $\textbf{w}_n$, there are some correlated filters as observed in Fig.~\ref{fig:conv1_lda}. 
During the subsequent retraining with our regularizer, regularization gradients coordinate a cluster of filters closer to a typical direction $\textbf{d}_m~(m \in [1...M], M \ll N) $, but data loss gradients avoid collapsing $\textbf{d}_m$ together so as to maintain the filters' capability of extracting discriminative features. 
If all filters could be extremely collapsed toward one point meanwhile maintain classification accuracy, it implies the filters are over-redundant and we can attain a very efficient DNN by decomposing it to a rank-one space.

We derive the \textit{Force Regularization} gradient from the \textit{normalized} filters based on the following facts:
(1)~A normalized filter is on the unit hypersphere, and its orientation is the only free parameter we need to optimize;
(2)~The gradient of $\mathbf{W}_{i}$ can be easily scaled by the vector length $ ||\mathbf{W}_{i}||$ without changing the angular velocity.

In Eq.~(\ref{eq:simi_reg}), $\textbf{f}_{ji} = f(\mathbf{w}_j - \mathbf{w}_i)$ is the force function related to distance. We study \textit{$\ell_2$-norm Force }
\begin{equation}
\label{eq:l2_force}
f_{\ell_2}(\mathbf{w}_j - \mathbf{w}_i) = \mathbf{w}_j - \mathbf{w}_i
\end{equation}
and \textit{$\ell_1$-norm Force}
\begin{equation}
\label{eq:l1_force}
f_{\ell_1}(\mathbf{w}_j - \mathbf{w}_i) = \frac{\mathbf{w}_j - \mathbf{w}_i}{||\mathbf{w}_j - \mathbf{w}_i||}
\end{equation}
in this work.
We define the force of Eq.~(\ref{eq:l2_force}) as \textit{$\ell_2$-norm Force} because the strength linearly decreases with the distance $||\mathbf{w}_j - \mathbf{w}_i||$, just as the gradient of regularization $\ell_2$-norm does. 
We name the force of Eq.~(\ref{eq:l1_force}) as \textit{$\ell_1$-norm Force} because the gradient is a constant unit vector regardless of the distance, just as the gradient of sparsity regularization $\ell_1$-norm is.

%
\subsection{Mathematical Implications}
\label{method:math}

\begin{table}
  \caption{ Ranks \textit{vs.} scalers of step sizes of regularization gradients. }
  \label{tab:cifar10_full:scaler}
  \centering
  	\begin{tabular}{cllll}
  		\toprule
  		Scaler & Error & conv1\textsuperscript{*} & conv2 & conv3 \\
  		\midrule
  		0 (baseline) & 18.0\%
  		& 17/32
  		& 27/32
  		& 55/64 \\
  		
  		$||\mathbf{W}_i||$ & 17.9\%
  		& 15/32
  		& 22/32
  		& 30/64 \\
  		
  		$ {1} / {||\mathbf{W}_i||}$ & 18.0\%
  		& 16/32
  		& 27/32
  		& 32/64 \\
  		\bottomrule
  		\multicolumn{5}{l}{\textsuperscript{*} {\small The first convolutional layer.}}
  	\end{tabular}
\end{table} 
\begin{table*}[b]
  \caption{ The \textit{rank M} in each convolutional layer after \textit{Force Regularization}.}
  \label{tab:ranks}
  \centering
  \resizebox{.85\textwidth}{!}{
  	\begin{tabular}{ccccccccc}
  		\toprule
  		Net & Force & Top-1 error & conv1 & conv2 & conv3 & conv4 & conv5 & Average rank ratio \textsuperscript{ $\ddagger$ }\\
  		\midrule
  		\textit{ConvNet} &
  		None (baseline)\textsuperscript{$\dagger$} & 18.0\%
  		& 17/32\textsuperscript{$\ddagger$} & 27/32 & 55/64 & -- & -- & 74.48\% \\
  		
  		\textit{ConvNet} &
  		$\ell_2$-norm & 17.9\%
  		& 15/32 & 22/32 & 30/64 & -- & -- & 54.17\% \\
  		
  		\textit{ConvNet} &
  		$\ell_1$-norm & 18.0\%
  		& 17/32 & 25/32 & 20/64 & -- & -- & 54.17\% \\
  		\cmidrule{1-9}
  		
  		\textit{AlexNet} &
  		None (baseline) & 42.63\%
  		& 47/96 & 164/256 & 306/384 & 318/384 & 220/256 & 72.29\% \\
  		
  		\textit{AlexNet} &
  		$\ell_2$-norm & 42.70\%
  		& 49/96 & 143/256 & 128/384 & 122/384 & 161/256 & 46.98\% \\
  		
  		\textit{AlexNet} &
  		$\ell_1$-norm & 42.45\%
  		& 49/96 & 155/256 & 157/384 & 108/384 & 178/256 & 50.03\% \\
  		
  		\bottomrule
  		\multicolumn{9}{l}{\textsuperscript{$\dagger$}The baseline without \textit{Force Regularization}. \textsuperscript{$\ddagger$}$M$/$N$: Low \textit{rank} $M$ over full rank $N$, which is defined as rank ratio.} 
 		
  	\end{tabular}
  }
\end{table*} 

This section explains the mathematical implications behind: 
\textit{Force Regularization} is related to but \textit{different} from minimizing the sum of pair-wise distances between normalized filters.
\begin{theorem}
\label{theo:eq_reg_l2_norm}
Suppose filter $\mathcal{W}_n \in \mathcal{W}$ is reshaped as a vector $\mathbf{W}_{n}\in\mathbb{R}^{1 \times CHW}$ and normalized as $\mathbf{w}_{n}\in\mathbb{R}^{1 \times CHW}(\forall n\in [1...N])$. For each filter, \textit{Force Regularization} under \textit{$\ell_2$-norm force} has the same gradient direction of regularization $R(\mathcal{W})$, but differs by adapting the step size to the filer's length, where
\begin{equation}
R(\mathcal{W}) = \frac{1}{2} \sum_{j=1}^{N} \sum_{i=1}^{N} \left|\left| \frac{\mathbf{W}_j}{||\mathbf{W}_j||} - \frac{\mathbf{W}_i}{||\mathbf{W}_i||} \right|\right|^2.
\label{eq:l2_norm_force_reg}
\end{equation}
\end{theorem}

\textit{Proof}:
Because $\mathbf{w}_j = \frac{\mathbf{W}_j}{||\mathbf{W}_j||}$,
\begin{equation}
\begin{split}
\frac{\partial R(\mathcal{W})}{\partial \mathbf{W}_i}
& = \frac{1}{2}  \sum_{j=1}^{N} \frac{ \partial \left( \mathbf{w}_j - \mathbf{w}_i \right) \left( \mathbf{w}_j - \mathbf{w}_i \right) ^ T }{\partial \mathbf{W}_i} \\
& = \frac{1}{2}  \sum_{j=1}^{N} \frac{ \partial \left( 1 - 2 \mathbf{w}_j  \mathbf{w}^T_i + 1 \right) }{\partial \mathbf{W}_i} \\
& = - \sum_{j=1}^{N} \frac{ \partial \left( \mathbf{w}_j  \mathbf{w}^T_i  \right) }{\partial \mathbf{W}_i}
= - \sum_{j=1}^{N}  \mathbf{w}_j \frac{ \partial \mathbf{w}^T_i }{\partial \mathbf{W}_i},
\end{split}
\label{eq:l2_norm_proof1}
\end{equation}
where $\frac{ \partial \mathbf{w}^T_i }{\partial \mathbf{W}_i} := \mathbf{G}_i$ is a derivative matrix with element
\begin{equation}
\begin{split}
G^{(pq)}_{i}
& = \frac{\partial w^{(p)}_{i}}{\partial W^{(q)}_{i}}
=  \frac{\partial \frac{W^{(p)}_{i}}{||\mathbf{W}_i||}}{\partial W^{(q)}_{i}} \\
& = \frac{1}{||\mathbf{W}_i||} \left( \delta(p,q) - \frac{W^{(p)}_{i} W^{(q)}_{i} }{||\mathbf{W}_i||^2}  \right).
\end{split}
\label{eq:l2_norm_proof2}
\end{equation}

Superscripts $p,q \in [1\ldots CHW]$ index the elements in vectors $\mathbf{w}_{i}$ and $\mathbf{W}_{i}$. $\delta (p,q)$ is the \textit{unit impulse} function:
\begin{equation}
\delta(p,q)=\begin{cases}
1~~~p=q\\
0~~~p \neq q
\end{cases}.
\end{equation}
Therefore,
\begin{equation}
\begin{split}
\mathbf{G}_{i}
& = \frac{1}{||\mathbf{W}_i||} \left( \mathbf{I} - \mathbf{w}^T_{i} \mathbf{w}_{i} \right).
\end{split}
\label{eq:l2_norm_proof3}
\end{equation}
Replacing Eq.~(\ref{eq:l2_norm_proof3}) to Eq.~(\ref{eq:l2_norm_proof1}), we have
\begin{equation}
\begin{split}
- \frac{\partial R(\mathcal{W})}{\partial \mathbf{W}_i}
& = \frac{1}{||\mathbf{W}_i||} \sum_{j=1}^{N} \left(  \left( \mathbf{w}_j - \mathbf{w}_i \right) - \left( \mathbf{w}_j - \mathbf{w}_i \right)  \mathbf{w}^T_{i} \mathbf{w}_{i} \right) \\
& = \frac{1}{||\mathbf{W}_i||}  \left( \left(\sum_{j=1}^{N} \mathbf{f}_{ji} \right) - \left(\sum_{j=1}^{N} \mathbf{f}_{ji} \right)  \mathbf{w}^T_i  \mathbf{w}_{i} \right),
\end{split}
\label{eq:l2_norm_proof_final}
\end{equation}
where $\mathbf{f}_{ji} = f_{\ell_2}(\mathbf{w}_j - \mathbf{w}_i) = \mathbf{w}_j - \mathbf{w}_i $. 
Therefore, Eq.~(\ref{eq:l2_norm_proof_final}) and Eq.~(\ref{eq:simi_reg}) have the same direction.

Theorem~\ref{theo:eq_reg_l2_norm} states that our proposed \textit{Force Regularization} in Eq.~(\ref{eq:simi_reg}) is related to Eq.~(\ref{eq:l2_norm_proof_final}). 
However, the step size of the gradient in Eq.~(\ref{eq:simi_reg}) is scaled by the length $||\mathbf{W}_i||$ of the filter instead of its reciprocal in Eq.~(\ref{eq:l2_norm_proof_final}). 
This ensures that the filter spins the same angle regardless of its length and avoids the issue of being divided by zero. 
Table~\ref{tab:cifar10_full:scaler} summarizes the ranks \textit{vs.} step sizes for the \textit{ConvNet} \cite{Alex_NIPS2012_4824}, which is trained by CIFAR-10 database without data augmentation. The original \textit{ConvNet} has $32$, $32$, and $64$ filters in each convolutional layer, respectively. The rank is the smallest number of basis filters (in~Fig.~\ref{fig:lra_3d_filters}) obtained by PCA with $\leq 5\%$ reconstruction error. 
Therefore, $||\mathbf{W}_i||$ works better than its reciprocal when coordinating filters to a lower-rank space.

%

Following the same proof procedure, we can easily find that \textit{Force Regularization} under \textit{$\ell_1$-norm Force} has the same conclusion when
\begin{equation}
R(\mathcal{W}) = \sum_{j=1}^{N} \sum_{i=1}^{N} \left|\left| \frac{\mathbf{W}_j}{||\mathbf{W}_j||} - \frac{\mathbf{W}_i}{||\mathbf{W}_i||} \right|\right|.
\label{eq:l1_norm_force_reg}
\end{equation}


\section{Experiments}
\label{sec:exp}
\subsection{Implementation}
\label{method:impl}
Our experiments are performed in Caffe~\cite{Caffe_2014} using CIFAR-10 \cite{krizhevsky2009learning} and ILSVRC-2012 ImageNet \cite{deng2009imagenet}. 
Published models are adopted as the baselines:
In CIFAR-10, we choose \textit{ConvNet} without data augmentation~\cite{Alex_NIPS2012_4824} and \textit{ResNets-20} with data augmentation~\cite{Kaiming_ResNet_ICCV}. 
We adopt the same shortcut connections in \cite{Wen_NIPS2016} for \textit{ResNets-20}.
For ImageNet, we use \textit{AlexNet} and \textit{GoogLeNet} models trained by Caffe, and report accuracy using only center crop of images.

Our experiments of \textit{Force Regularization} show that, with the same maximum iterations, the training from the baseline can achieve a better tradeoff between accuracy and speedup comparing with the training from scratch, because the baseline offers a good initial point for both accuracy and filter correlation. 
During the training with \textit{Force Regularization} on CIFAR-10, we use the same base learning rate as the baseline; 
while in ImageNet, $0.1\times$ base learning rate of the baseline is adopted.

\subsection{Rank Analysis of Coordinated DNNs}
\label{method:exp_lowrank}

In light of various low-rank approximation methods, without losing the generalization, we first adopt \textit{Principal Component Analysis} (PCA)~\cite{7332968}\cite{Liu_CVPR2015} to evaluate the effectiveness of \textit{Force Regularization}. 
Specifically, the filter tensor $\mathcal{W}$ can be reshaped to a matrix $\mathbf{W} \in \mathbb{R}^{N \times CHW} $, the rows of which are the reshaped filters $\mathbf{W}_n~(\forall n \in [1...N])$.
PCA minimizes the \textit{least square reconstruction error} when projecting a column $(\mathbb{R}^N)$ of $\mathbf{W}$ to a low-rank space $\mathbb{R}^M (M \ll N)$.
The reconstruction error is $e_M = \sum_{i=M+1}^{N} \lambda_i$, where $\lambda_i$ is the $i$-th largest eigenvalue of covariance matrix $\frac{\mathbf{W}  \mathbf{W}^T}{CHW-1}$.
Under the constraint of \textit{error percentage} $\frac{ e_M }{ e_0 }$ (\textit{e.g.}, $\frac{ e_M }{ e_0 } \leq 5\%$), \textit{lower}-rank approximation can be obtained if the minimal rank $M$ can be \textit{smaller}.  
In this section, without explicit explanation, we define \textit{rank} $M$ of a convolutional layer as the minimal $M$ which has $\leq 5\%$  reconstruction error by PCA.

\begin{figure}
	\centering
	\includegraphics[width=1.0\columnwidth]{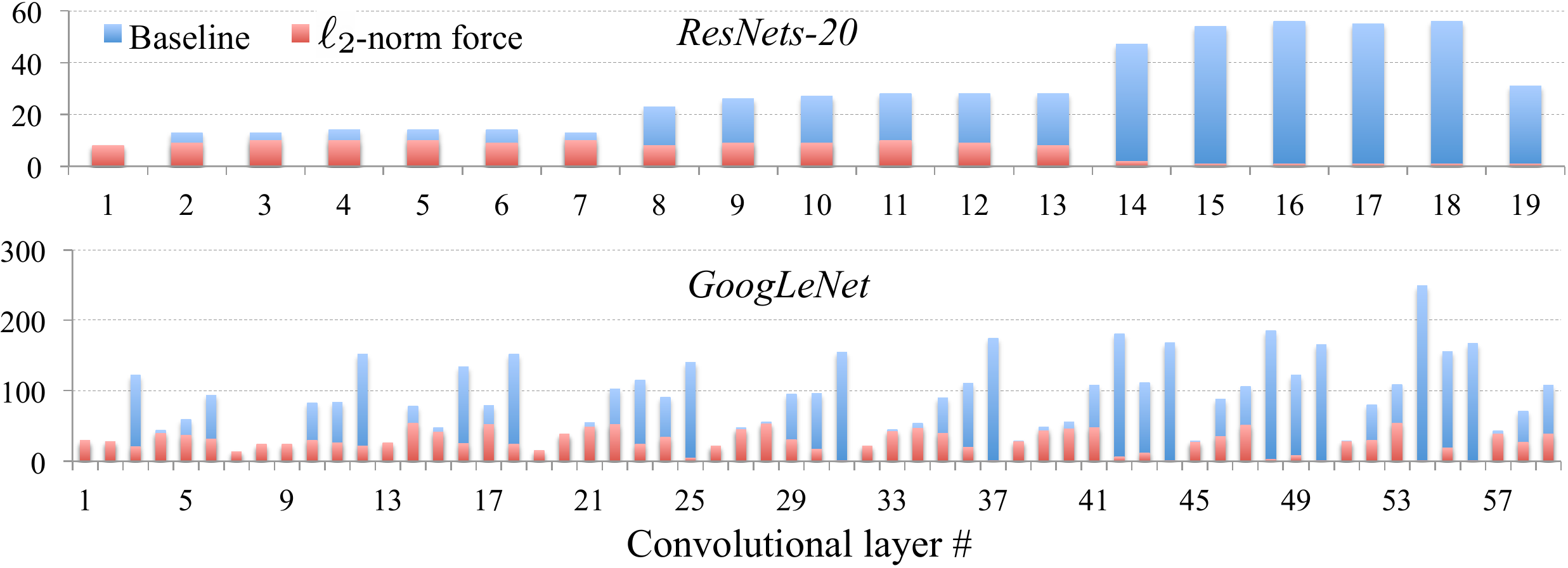}
	
	\caption{The \textit{rank}~$M$ in each convolutional layer of \textit{ResNets-20} and \textit{GoogLeNet}. Red bar overlaps blue bar. The accuracy loss is 0.75\% for \textit{ResNets-20} and 2.46\% (top-5) for \textit{GoogLeNet}.}
	\label{fig:rankratio_resnet}
\end{figure}

Table~\ref{tab:ranks} summarizes the \textit{rank} $M$ in each layer of \textit{ConvNet} and \textit{AlexNet} without accuracy loss after \textit{Force Regularization}. 
In the baselines, the learned filters in the front layers are intrinsically in a very low-rank space but the \textit{rank} $M$ in deeper layers is high. 
This could explain why only speedups of the first two convolutional layers were reported in~\cite{Denton_NIPS2014}.
Fortunately, by using either $\ell_2$-norm or $\ell_1$-norm force, our method can efficiently maintain the low \textit{rank} $M$ in the first two layers (\textit{e.g.}, conv1-conv2 in \textit{AlexNet}), meanwhile significantly reduce the \textit{rank} $M$ of deeper layers (\textit{e.g.}, conv3-conv5 in \textit{AlexNet}).
On average, our method can reduce the layer-wise rank ratio by $\sim50\%$.
The effectiveness of our method on deep layers is very important as the depth of modern DNNs grows dramatically~\cite{GoogleNet_2015}\cite{Kaiming_ResNet_ICCV}. 
Fig.~\ref{fig:rankratio_resnet} shows the \textit{rank}~$M$ of \textit{ResNets-20}~\cite{Kaiming_ResNet_ICCV} and \textit{GoogLeNet}~\cite{GoogleNet_2015} after \textit{Force Regularization}, representing the scalability of our method on deeper DNNs. 
With an acceptable accuracy loss, $5$ layers in \textit{ResNets-20} and $6$ layers in \textit{GoogLeNet} are even coordinated to \textit{rank} $M=1$, which indicates those Inception blocks in \textit{GoogLeNet} or Residual blocks in \textit{ResNets} have been over-parameterized and can be greatly simplified.

\begin{figure}
	\centering
	\includegraphics[width=.9\columnwidth]{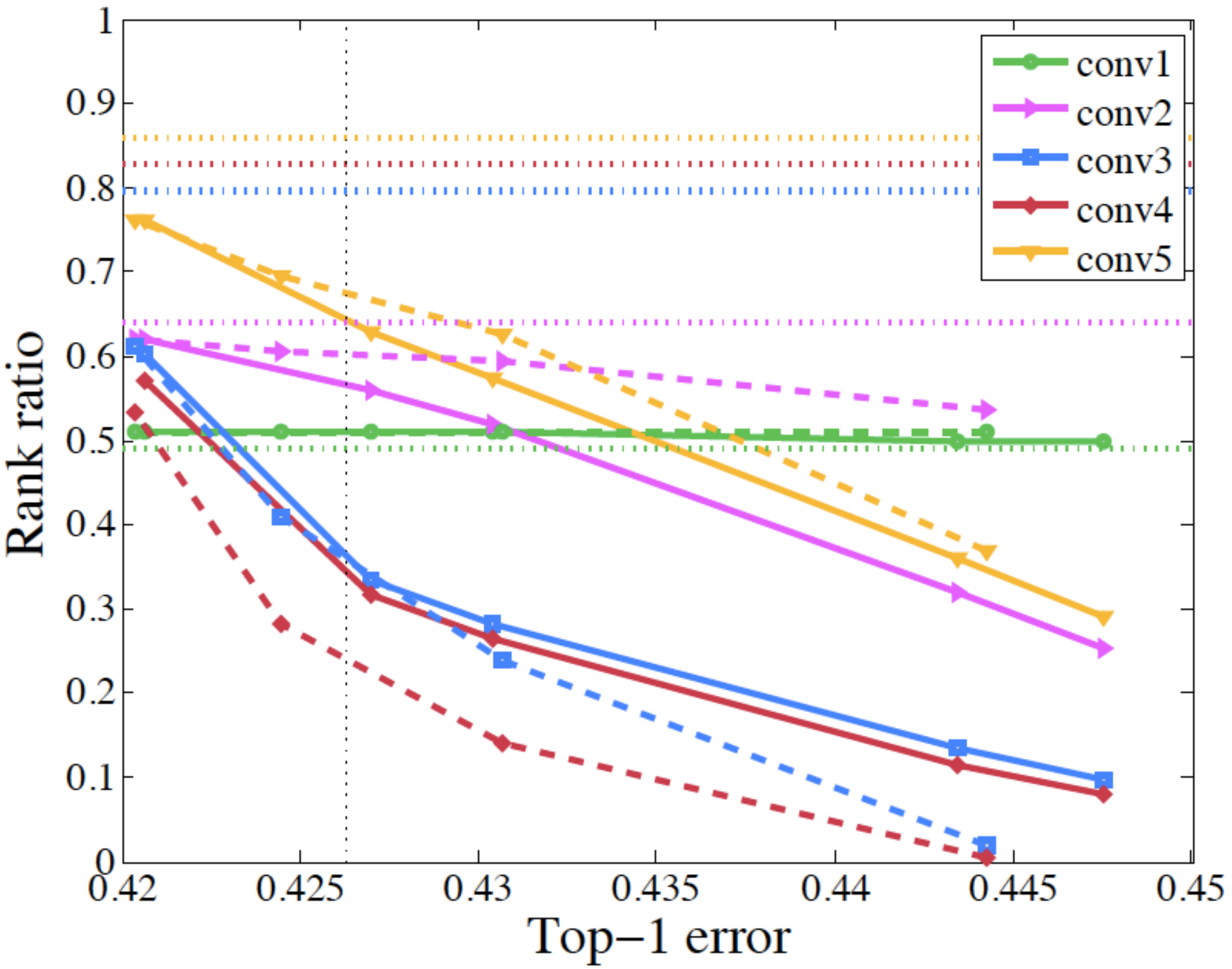}
	\caption{Th rank ratio (having $\leq 5\%$ PCA reconstruction error) in each layer \textit{vs.} top-1 error for \textit{AlexNet}. Horizontal dotted lines represent the rank ratios of the baseline, and vertical dotted line is the error of baseline. Solid (dashed) curves depict rank ratios of the \textit{AlexNet} after \textit{Force Regularization} by $\ell_2$-norm ($\ell_1$-norm) force. Each layer is denoted by a typical color. The sensitivity of hyper-parameter $\lambda_s$: along the direction from left to right, $\lambda_s$ of $\ell_2$-norm force changes from 1.2e-5, to 1.8e-5, 2.0e-5, 3.0e-5, and 3.5e-5; and for $\ell_1$-norm force, it changes from 1.5e-5, to 1.8e-5, 2.0e-5, and 2.5e-5.}
	\label{fig:rank_ratio_vs_error}
\end{figure}


To study the trade-off between rank, accuracy, and the pros and cons of $\ell_2$-norm and $\ell_1$-norm force, we conducted comprehensive experiments on \textit{AlexNet}. 
As shown in Fig.~\ref{fig:rank_ratio_vs_error}, with mere 1.71\% (1.80\%) accuracy loss, the average rank ratio can be reduced to 28.59\% (28.72\%) using $\ell_2$-norm ($\ell_1$-norm) force. 
Very impressively, the \textit{rank}~$M$  of each group in conv4 can be reduced to one by $\ell_1$-norm force. 
The results also show that $\ell_2$-norm force is more effective than $\ell_1$-norm force when the rank ratio is high (\textit{e.g.}, conv2 and conv5), while $\ell_1$-norm force works better for layers whose potential rank ratios are low (\textit{e.g.}, conv3 and conv4). 
In general, $\ell_2$-norm force can better balance the ranks across all the layers.

%


\begin{figure}
	\centering
	\includegraphics[width=1.0\columnwidth]{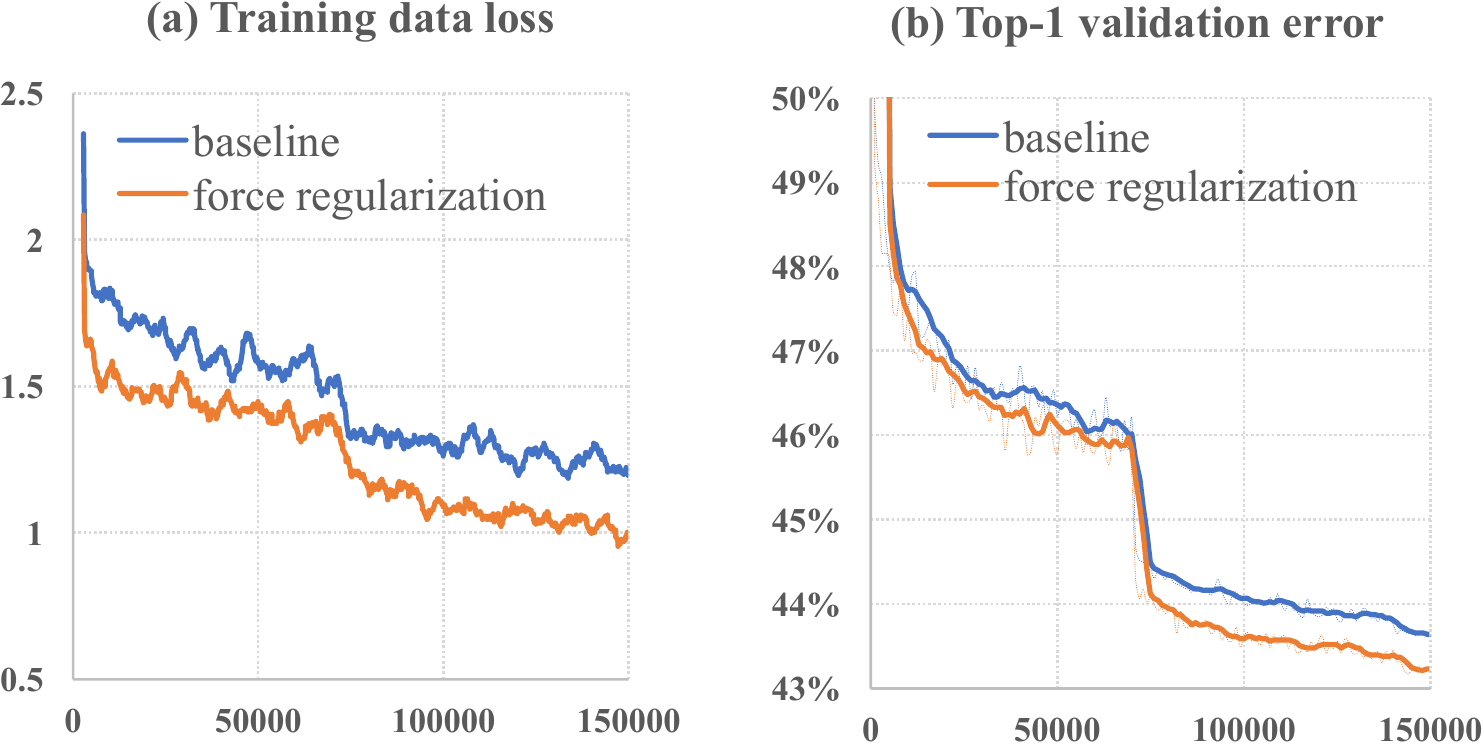}
	\caption{Training data loss and top-1 validation error vs. iteration when fine-tuning \textit{AlexNet} which is decomposed to the same ranks.}
	\label{fig:convergence}
\end{figure} 

Because \textit{Force Regularization} coordinates more useful weight information in a low-rank space, it essentially can provide a better training initialization for the DNNs that are decomposed by LRA. 
Fig.~\ref{fig:convergence} plots the training data loss and top-1 validation error of \textit{AlexNet}, which is decomposed to the same ranks by PCA. 
The baseline is the original \textit{AlexNet} and the other \textit{AlexNet} is coordinated by \textit{Force Regularization}. 
The figure shows that the error sharply converges to a low level after a few iterations, indicating LRA provides a very good initialization for the low-rank DNNs. 
Training it from scratch has significant accuracy loss.
More importantly, DNNs coordinated by \textit{Force Regularization} can converge faster to a lower error. 

\begin{table}[b]
  \caption{  The accuracy of different LRA under the same ranks. }
  \label{tab:pca_svd_kmeans}
  \centering
  \resizebox{.8\columnwidth}{!}{
  	\begin{tabular}{ccc}
  		\toprule
  		\textit{Force} & LRA &  Top-1 error   \\
  		\midrule 
  		
  		\multirow{3}{*}{None} & PCA  &43.21\%  \\
  		& SVD\textsuperscript{$\dagger$}  &43.27\%  \\
  		& k-means\textsuperscript{$\dagger$}  &44.34\%  \\
  		
  		\midrule 
  		
  		\multirow{3}{*}{$\ell_2$-norm} & PCA  &43.25\%  \\
  		& SVD\textsuperscript{$\dagger$}  &  43.20\%  \\
  		& k-means\textsuperscript{$\dagger$}  & 44.80\%  \\

  		\bottomrule
  		\multicolumn{3}{l}{\textsuperscript{$\dagger$} SVD and k-means preserve the same ranks with PCA }
  	\end{tabular}
  }
\end{table} 

Besides PCA~\cite{Liu_CVPR2015}\cite{7332968}, we also evaluated the effectiveness of \textit{Force Regularization} when integrating it with SVD~\cite{Denton_NIPS2014}\cite{Cheng_arxiv_2015} or k-means clustering~\cite{Denton_NIPS2014}\cite{bauckhage2015kmeans}. 
Table~\ref{tab:pca_svd_kmeans} compares the accuracies of \textit{AlexNet} decomposed by different LRA methods. 
All LRAs preserve the same ranks in all layers, which means the decomposed \textit{AlexNet} have the same network structure. 
In summary, PCA and SVD obtain similar accuracy and surpass k-means clustering. 
Due to the limited pages, we adopt PCA as the representative in our study.

\subsection{Acceleration of DNN Testing}
\label{method:exp_speedup}

In our experiments, we first train DNNs with \textit{Force Regularization}, then decompose DNNs using LRA methods and fine-tune them to recover accuracy.
In evaluation of speed, we omit small CIFAR-10 database and focus on large-scale DNNs on ImageNet, whose speed is a real concern.
To prove the effective acceleration of \textit{Force Regularization}, we adopt the speedup of state-of-the-art LRAs~\cite{7332968}\cite{Denil_NIPS2013_5025}\cite{Cheng_arxiv_2015} as our baseline. 
Our speedup is achieved in the case that the DNN filters are first coordinated by \textit{Force Regularization} and then decomposed using the same LRAs. 
The practical GPU speed is profiled by the advanced hardware (NVIDIA GTX 1080) and software (cuDNN 5.0).
The CPU speed is measured in Intel Xeon E5-2630 and ATLAS library.
The batch size is 256.


\textbf{Cross-filter LRA:}~
We first evaluate the speedup of cross-filter LRA shown in Fig.~\ref{fig:lra_3d_filters}.
In previous works~\cite{Denton_NIPS2014}\cite{Cheng_arxiv_2015}, the optimal rank in each layer can be selected layer-by-layer using cross validation. 
However, the number of hyper-parameters increases linearly with the depth of DNNs. 
To save development time, we utilize an identical \textit{error percentage} $\frac{e_M}{e_0}$ across all layers as the single hyper-parameter although layer-wise rank selection may give better tradeoff. 
The rank in a layer is the minimal $M$ which has error $ \leq \frac{e_M}{e_0}$.

As aforementioned in Section~\ref{method:exp_lowrank} and Table~\ref{tab:ranks}, the learned conv1 and conv2 of \textit{AlexNet} are already in a very low-rank space and achieve good speedups using LRAs~\cite{Denton_NIPS2014}. 
Thus we mainly focus on conv3-conv5 here. 
Table~\ref{tab:speedups} summarizes the speedups of PCA approximation of \textit{AlexNet} with and without $\ell_2$-norm \textit{Force Regularization}.
With ignoble accuracy difference, \textit{Force Regularization} successfully coordinates filters to a lower-rank space and accelerates the testing by a higher factor, comparing with the state-of-the-art LRA.
Similar results are observed when applying $\ell_1$-norm force.

Results in Table~\ref{tab:speedups} also show that practical speedup is different from theoretical speedup. 
Generally, the difference 
is smaller in lower-performance processors. 
In CPU mode of Table~\ref{tab:speedups}, \textit{Force Regularization} achieves $2 \times$ speedup of total convolutional time.

\begin{table}
	\caption{ The higher speedups of \textit{AlexNet} by \textit{Force Regularization}.}
	\label{tab:speedups}
	\centering
	\resizebox{\columnwidth}{!}{
	\begin{tabular}{cc|rrrr}
		\toprule [0.12 em]
		Force & Top-1 error & &  conv3 & conv4 & conv5   \\
		\midrule [0.12 em]
		
		None & 43.21\% & rank& 184 & 201 & 146  \\
		$\ell_2$-norm & 43.25\% & rank & 124 & 106 & 129  \\
		\cmidrule{1-6}
		None & 43.21\% & GPU & 1.58$\times$ & 1.21$\times$ & 1.15$\times$\\
		$\ell_2$-norm & 43.25\% & GPU & 2.16$\times$ & 2.03$\times$ & 1.33$\times$ \\
		\cmidrule{1-6}
		None & 43.21\% & CPU & 1.78$\times$ & 1.60$\times$ & 1.47$\times$\\
		$\ell_2$-norm & 43.25\% & CPU & 2.45$\times$ & 2.76$\times$ & 1.64$\times$ \\
		\cmidrule{1-6}
		None & 43.21\% & theoretical & 1.79$\times$& 1.72$\times$ & 1.63$\times$ \\
		$\ell_2$-norm & 43.25\% & theoretical & 2.65$\times$ & 3.26$\times$ & 1.85$\times$ \\
		
		\bottomrule [0.12 em]
	\end{tabular}
	}
\end{table}

\textbf{Speeding up state-of-the-art LRA}:~
We also duplicate the state-of-the-art work \cite{Cheng_arxiv_2015} as the baseline\footnote{Code is provided by the authors in \url{https://github.com/chengtaipu/lowrankcnn/}} ($lra_1$). 
After LRA, \textit{AlexNet} is fine-tuned with learning rate starting from 0.001 and divided by 10 at iteration 70,000 and 140,000. Fine-tuning terminates after 150,000 iterations.

The first row in Table~\ref{tab:iclr_speedups} contains the results of the baseline~\cite{Cheng_arxiv_2015}, which don't scale well to the advanced ``TITAN 1080 + cuDNN 5.0'' in conv3--5. 
This is because $3\times3$ convolution is highly optimized in cuDNN 5.0, \textit{e.g.}, using Winograd's minimal filtering algorithms~\cite{winograd2015fast}. 
However, the baseline decomposes the $3\times3$ convolution to a pair of $3\times1$ and $1\times3$ convolution so that the optimized cuDNN is not fully exploited. 
This will be a common issue in the baseline, considering Winograd's algorithm is universally used and $3\times3$ convolution is one of the most common structures.
We find that LRA in Fig.~\ref{fig:lra_3d_filters} can be utilized for conv 3--5 to solve this issue, because it can maintain the $3\times3$ shape. 
We name this LRA as $lra_2$, which decomposes conv1--conv2 using LRA in~\cite{Cheng_arxiv_2015} and conv 3--5 using LRA of Fig.~\ref{fig:lra_3d_filters}.
The second row in Table~\ref{tab:iclr_speedups} shows that our  $lra_2$ can scale well to the hardware and software advances of ``TITAN 1080 + cuDNN 5.0''.
More importantly, \textit{Force Regularization} on conv3--5 can enforce them to more lightweight layers and attain higher speedup factors than $lra_2$ without using it. 
The result is shown in the third row, which in total achieves $2.03\times$ speedup for the whole convolution in GPU. 
With small accuracy loss in row 4 of Table~\ref{tab:iclr_speedups}, \textit{Force Regularization} achieves $2.50\times$ speedup of total convolution on GPU and $4.05 \times$ on CPU.

\begin{table}
	\caption{ The higher speedup factors by force regularization.}
	\label{tab:iclr_speedups}
	\centering
	\resizebox{\columnwidth}{!}{
	\begin{tabular}{ccc|rccc}
		\toprule [0.12 em]
		LRA & Force & Top-5 err. & & conv3 & conv4 & conv5   \\
		\midrule [0.12 em]
		
		\textit{$lra_1$}~\cite{Cheng_arxiv_2015} & None & 20.65\%  & GPU  & 0.86$\times$ & 0.57$\times$ & 0.40$\times$\\
		
		
		
		\cmidrule{1-7}
		\textit{$lra_2$} & None & 19.93\% & GPU  & 1.89$\times$ & 1.57$\times$ & 1.57$\times$\\
		
		\cmidrule{1-7}
		\textit{$lra_2$} & $\ell_2$-norm & 20.14\%  & GPU  & 2.25$\times$ & 2.03$\times$ & 1.60$\times$\\
		
		\bottomrule [0.12 em]
		\multirow{2}{*}{\textit{$lra_2$}} & \multirow{2}{*}{$\ell_2$-norm} & \multirow{2}{*}{21.68\%} 
		& GPU  & 3.56$\times$ & 3.01$\times$ & 2.40$\times$\\
		&&& CPU  & 4.81$\times$ & 4.00$\times$ & 2.92$\times$\\
		
		\bottomrule [0.12 em]
	\end{tabular}
	}
\end{table}

\begin{table*}[t]
	\caption{ Comparison of speedup factor on \textit{AlexNet} by state-of-the-art DNN acceleration methods.}
	\label{tab:speedup_summary}
	\centering
	\resizebox{.82\textwidth}{!}{
	\begin{tabular}{cc|ccccc|c}
		\toprule [0.12 em]
		Method  & Top-5 err. & conv1 & conv2 & conv3 & conv4 & conv5 & total  \\
		\midrule [0.12 em]
		\textit{AlexNet in Caffe}  & 19.97\%  &  1.00$\times$  &  1.00$\times$  & 1.00$\times$  & 1.00$\times$ & 1.00$\times$ & \textbf{1.00}$\times$ \\
		
		\cmidrule{1-8}
		\textit{cp-decomposition}~\cite{lebedev2014speeding} & 20.97\% (+1.00\%)  &  --  &  4.00$\times$  & -- & -- & -- &  \textbf{1.27}$\times$ \\
		
		\cmidrule{1-8}
		\textit{one-shot}~\cite{kim2015compression} & 21.67\% (+1.70\%)  &  1.48$\times$  &  2.30$\times$  & 3.84$\times$ & 3.53$\times$ & 3.13$\times$ & \textbf{2.52}$\times$ \\

		\cmidrule{1-8}
		\multirow{2}{*}{\textit{SSL}~\cite{Wen_NIPS2016}} 
		& 19.58\% (-0.39\%) &  1.00$\times$  &  1.27$\times$  & 1.64$\times$ & 1.68$\times$ & 1.32$\times$ & \textbf{1.35}$\times$\\
		& 21.63\% (+1.66\%) &  1.05$\times$  &  3.37$\times$  & 6.27$\times$ & 9.73$\times$ & 4.93$\times$ &  \textbf{3.13}$\times$\\

		\cmidrule{1-8}
		\multirow{2}{*}{\textit{our} $lra_2$} & 20.14\% (+0.17\%)  &  2.61$\times$  &  6.06$\times$  & 2.48$\times$ & 2.20$\times$ & 1.58$\times$ & \textbf{2.69}$\times$ \\
		 & 21.68\% (+1.71\%)  &  2.65$\times$  &  6.22$\times$  & 4.81$\times$ & 4.00$\times$ & 2.92$\times$ & \textbf{4.05}$\times$\\

		\bottomrule [0.12 em]
		
	\end{tabular}
	}
\end{table*}

Table~\ref{tab:speedup_summary} compares our method with state-of-the-art DNN acceleration methods, in CPU mode. When the speedup of total time was not reported by the authors, we estimate it by the weighted average speedups over all layers, where the weighting coefficients are derived from the percentage of running time of each layer. In our hardware platform, conv1--conv5 respectively consume $15.89\%$, $28.25\%$, $24.32\%$, $18.70\%$ and $12.84\%$ testing time. The estimation is accurate, for example, we estimate $2.58\times$ of total time in \textit{one-shot}~\cite{kim2015compression}, which is very close to $2.52\times$ reported by the authors.
Comparing with both \textit{cp-decomposition} and \textit{one-shot} methods, our method can achieve higher accuracy and higher speedup.
Comparing with \textit{SSL}, with almost the same top-5 error ($21.68\%$ vs. $21.63\%$), we can attain higher speedup of $4.05\times$ \textit{vs.} $3.13\times$.

\textit{deep-compression}~\cite{Han_ICLR2016} reported $3\times$ to $4\times$ speedups in fully-connected layers when batch size was 1. However,  convolution is the bottleneck of DNNs, \textit{e.g.}, the convolution time in \textit{AlexNet} is $5\times$ of the time in fully-connected layers when profiled in our CPU platform. Moreover, no speedup was observed in the batching scenario as reported by the authors~\cite{Han_ICLR2016}.
More importantly, as we will show in Section~\ref{method:sparsity}, our work can work together with sparsity-based methods (\textit{e.g.}, \textit{SSL} or \textit{deep-compression}) to obtain \textit{lower-rank and sparse DNNs} and potentially further accelerate the testing of DNNs.

\subsection{Lower-rank and Sparse DNNs}
\label{method:sparsity}
\begin{figure}[b]
	\centering
	\includegraphics[width=1.0\columnwidth]{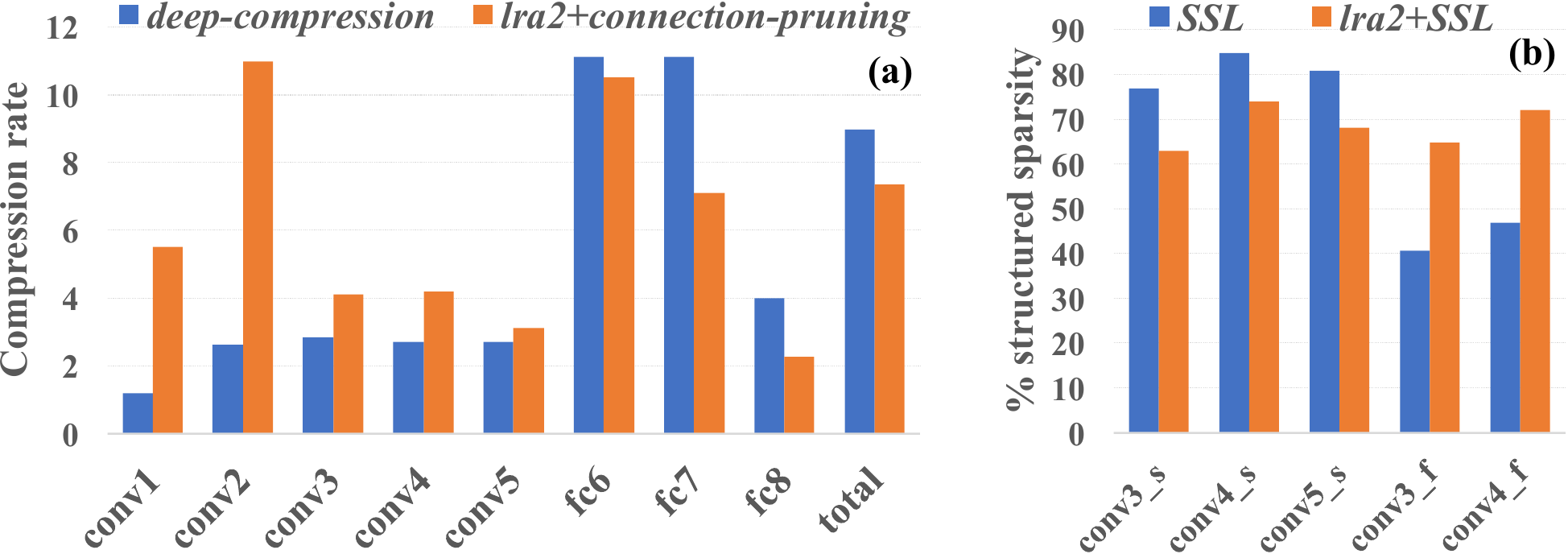}
	\caption{The results of sparsifying lightweight DNNs whose filters are coordinated to a lower-rank space by \textit{Force Regularization}. 
		In terms of \textit{deep-compression} in (a), we only count the compression rate obtained from connection pruning for a fair comparison, but quantization and Huffman coding can also be utilized to improve the compression rate for our model.
		Based on \textit{SSL} in (b), we enforce shape-wise sparsity on conv3\_s, conv4\_s and conv5\_s to learn the shapes of basis filters meanwhile enforce filter-wise sparsity on conv3\_f and conv4\_f to learn the number of filters~\cite{Wen_NIPS2016}.
		As each convolutional layer in the $lra_2$ is decomposed to two small layers, we respectively denote the first and second small layer by suffixing ``\_s'' and ``\_f''. 
		The baseline and our model have the same accuracy.}
	\label{fig:sparsity_comparison}
\end{figure} 

We sparsify the lightweight deep neural network (\textit{i.e.}, the first one of $lra_2$ in Table~\ref{tab:speedup_summary}), using Structured Sparsity Learning~\textit{SSL}~\cite{Wen_NIPS2016} or non-structured \textit{connection-pruning}~\cite{Jongsoo_ICLR2017}. 
Note that Guided Sparsity Learning (\textit{GSL}) is not adopted in our \textit{connection-pruning} though better sparsity is achievable when applying it. Figure~\ref{fig:sparsity_comparison} summarizes the results. 

Experiments prove that our method can work together with both structured and non-structured sparsity methods to further compress and accelerate models. 
Comparing with \textit{deep-compression} in Figure~\ref{fig:sparsity_comparison}(a), our model has comparable compression rates but $2.69\times$ faster testing time. 
Typically, our model has higher compression rates in convolutional layers, which provides more space for computation reduction and generalizes better to modern DNNs (\textit{ResNets-152}~\cite{Kaiming_ResNet_ICCV}, for example, whose parameters in fc layers are only $4\%$). 
In Figure~\ref{fig:sparsity_comparison}(b), our accelerated model can be further accelerated using \textit{SSL}. 
The shape-wise sparsity in conv3--5 of our model is slightly lower because our model is already aggressively compressed by LRA.
The higher filter-wise sparsity, however, implies the orthogonality of our approach to \textit{SSL}.

\subsection{Generalization of Force Regularization}
\label{method:accuracy}
\begin{table}[b]
  \caption{ Improved accuracy with \textit{Discrimination Regularization}. }
  \label{tab:discrimination}
  \centering
  	\begin{tabular}{ccc}
  		\toprule
  		\textit{Net} & Regularization & Top-1 error   \\
  		\midrule 
  		\textit{AlexNet} & None (baseline) & 42.63\% \\
  		\textit{AlexNet} & $\ell_2$-norm force & 41.71\% \\
  		\textit{AlexNet} & $\ell_1$-norm force & 41.53\% \\
  		
  		\cmidrule{1-3}
  		\textit{ResNets-20} & None (baseline) & 8.82\% \\
  		\textit{ResNets-20} & $\ell_2$-norm force & 7.97\% \\
  		\textit{ResNets-20} & $\ell_1$-norm force & 8.02\% \\

  		\bottomrule
  	\end{tabular}
\end{table}  

In convolutional layers, each filter basically extracts a discriminative feature, \textit{e.g.}, an orientation-selective pattern or a color blob in the first layer~\cite{Alex_NIPS2012_4824} or a high-level feature (\textit{e.g.}, textures, faces, \textit{etc.}) in deeper layers~\cite{zeiler2014visualizing}. 
The discrimination among filters is important for classification performance. 
Our method can coordinate filters for more lightweight DNNs meanwhile maintain the discrimination.
It can also be generalized to learn more discriminative filters to improve the accuracy.
The extension to \textit{Discrimination Regularization} is straightforward but effective: the opposite gradient of \textit{Force Regularization} (\textit{i.e.}, $\lambda_s<0$) is utilized to update the filter. In this scenario, it works as the \textit{repulsive force} to repel surrounding filters and enhance the discrimination.
Table~\ref{tab:discrimination} summarizes the improved accuracy of state-of-the-art DNNs.

\subsubsection*{Acknowledgments}
This work was supported in part by NSF CCF-1744082. Any opinions, findings and conclusions or recommendations expressed in this material are those of the authors and do not necessarily reflect the views of NSF or their contractors.

{\small
\bibliographystyle{ieee}
\bibliography{iccv2017}

\begin{thebibliography}{10}\itemsep=-1pt

\bibitem{alvarez2016learning}
J.~M. Alvarez and M.~Salzmann.
\newblock Learning the number of neurons in deep networks.
\newblock In {\em Advances in Neural Information Processing Systems (NIPS)},
  pages 2262--2270, 2016.

\bibitem{bauckhage2015kmeans}
C.~Bauckhage.
\newblock k-means clustering is matrix factorization.
\newblock {\em arXiv:1512.07548}, 2015.

\bibitem{deng2009imagenet}
J.~Deng, W.~Dong, R.~Socher, L.-J. Li, K.~Li, and L.~Fei-Fei.
\newblock Imagenet: A large-scale hierarchical image database.
\newblock In {\em IEEE Conference on Computer Vision and Pattern Recognition
  (CVPR)}, 2009.

\bibitem{Denil_NIPS2013_5025}
M.~Denil, B.~Shakibi, L.~Dinh, M.~A. Ranzato, and N.~de~Freitas.
\newblock Predicting parameters in deep learning.
\newblock In {\em Advances in Neural Information Processing Systems (NIPS)}.
  2013.

\bibitem{Denton_NIPS2014}
E.~L. Denton, W.~Zaremba, J.~Bruna, Y.~LeCun, and R.~Fergus.
\newblock Exploiting linear structure within convolutional networks for
  efficient evaluation.
\newblock In {\em Advances in Neural Information Processing Systems (NIPS)}.
  2014.

\bibitem{guo_dnf_nips2016}
Y.~Guo, A.~Yao, and Y.~Chen.
\newblock Dynamic network surgery for efficient dnns.
\newblock In {\em Advances in Neural Information Processing Systems (NIPS)}.
  2016.

\bibitem{Han_ICLR2016}
S.~Han, H.~Mao, and W.~J. Dally.
\newblock Deep compression: Compressing deep neural network with pruning,
  trained quantization and huffman coding.
\newblock {\em arXiv:1510.00149}, 2015.

\bibitem{Han_NIPS2015}
S.~Han, J.~Pool, J.~Tran, and W.~Dally.
\newblock Learning both weights and connections for efficient neural network.
\newblock In {\em Advances in Neural Information Processing Systems (NIPS)}.
  2015.

\bibitem{he2015delving}
K.~He, X.~Zhang, S.~Ren, and J.~Sun.
\newblock Delving deep into rectifiers: Surpassing human-level performance on
  imagenet classification.
\newblock In {\em International Conference on Computer Vision (ICCV)}, 2015.

\bibitem{Kaiming_ResNet_ICCV}
K.~He, X.~Zhang, S.~Ren, and J.~Sun.
\newblock Deep residual learning for image recognition.
\newblock In {\em IEEE Conference on Computer Vision and Pattern Recognition
  (CVPR)}, 2016.

\bibitem{Yani_arxiv_2015}
Y.~Ioannou, D.~P. Robertson, J.~Shotton, R.~Cipolla, and A.~Criminisi.
\newblock Training cnns with low-rank filters for efficient image
  classification.
\newblock {\em arXiv:1511.06744}, 2015.

\bibitem{Max_J_arxiv2014}
M.~Jaderberg, A.~Vedaldi, and A.~Zisserman.
\newblock Speeding up convolutional neural networks with low rank expansions.
\newblock In {\em Proceedings of the British Machine Vision Conference
  ({BMVC})}, 2014.

\bibitem{Caffe_2014}
Y.~Jia, E.~Shelhamer, J.~Donahue, S.~Karayev, J.~Long, R.~Girshick,
  S.~Guadarrama, and T.~Darrell.
\newblock Caffe: Convolutional architecture for fast feature embedding.
\newblock {\em arXiv:1408.5093}, 2014.

\bibitem{kim2015compression}
Y.-D. Kim, E.~Park, S.~Yoo, T.~Choi, L.~Yang, and D.~Shin.
\newblock Compression of deep convolutional neural networks for fast and low
  power mobile applications.
\newblock {\em arXiv:1511.06530}, 2015.

\bibitem{krizhevsky2009learning}
A.~Krizhevsky and G.~Hinton.
\newblock Learning multiple layers of features from tiny images.
\newblock 2009.

\bibitem{Alex_NIPS2012_4824}
A.~Krizhevsky, I.~Sutskever, and G.~E. Hinton.
\newblock Imagenet classification with deep convolutional neural networks.
\newblock In {\em Advances in Neural Information Processing Systems (NIPS)}.
  2012.

\bibitem{winograd2015fast}
A.~Lavin.
\newblock Fast algorithms for convolutional neural networks.
\newblock {\em arXiv:1509.09308}, 2015.

\bibitem{lebedev2014speeding}
V.~Lebedev, Y.~Ganin, M.~Rakhuba, I.~Oseledets, and V.~Lempitsky.
\newblock Speeding-up convolutional neural networks using fine-tuned
  cp-decomposition.
\newblock {\em arXiv:1412.6553}, 2014.

\bibitem{Lebedev_2016_CVPR}
V.~Lebedev and V.~Lempitsky.
\newblock Fast convnets using group-wise brain damage.
\newblock In {\em IEEE Conference on Computer Vision and Pattern Recognition
  (CVPR)}, 2016.

\bibitem{lecun1989optimal}
Y.~LeCun, J.~S. Denker, S.~A. Solla, R.~E. Howard, and L.~D. Jackel.
\newblock Optimal brain damage.
\newblock In {\em Advances in Neural Information Processing Systems (NIPS)},
  volume~2, pages 598--605, 1989.

\bibitem{Haoli_ICLR2017}
H.~Li, A.~Kadav, I.~Durdanovic, H.~Samet, and H.~P. Graf.
\newblock Pruning filters for efficient convnets.
\newblock In {\em International Conference on Learning Representations (ICLR)},
  2017.

\bibitem{Liu_CVPR2015}
B.~Liu, M.~Wang, H.~Foroosh, M.~Tappen, and M.~Pensky.
\newblock Sparse convolutional neural networks.
\newblock In {\em IEEE Conference on Computer Vision and Pattern Recognition
  (CVPR)}, 2015.

\bibitem{Jongsoo_ICLR2017}
J.~Park, S.~Li, W.~Wen, P.~T.~P. Tang, H.~Li, Y.~Chen, and P.~Dubey.
\newblock Faster cnns with direct sparse convolutions and guided pruning.
\newblock In {\em International Conference on Learning Representations (ICLR)},
  2017.

\bibitem{Vggnet_2014}
K.~Simonyan and A.~Zisserman.
\newblock Very deep convolutional networks for large-scale image recognition.
\newblock {\em arXiv:1409.1556}, 2014.

\bibitem{GoogleNet_2015}
C.~Szegedy, W.~Liu, Y.~Jia, P.~Sermanet, S.~Reed, D.~Anguelov, D.~Erhan,
  V.~Vanhoucke, and A.~Rabinovich.
\newblock Going deeper with convolutions.
\newblock In {\em IEEE Conference on Computer Vision and Pattern Recognition
  (CVPR)}, 2015.

\bibitem{Cheng_arxiv_2015}
C.~Tai, T.~Xiao, X.~Wang, and W.~E.
\newblock Convolutional neural networks with low-rank regularization.
\newblock In {\em International Conference on Learning Representations (ICLR)},
  2016.

\bibitem{wang2016accelerating}
P.~Wang and J.~Cheng.
\newblock Accelerating convolutional neural networks for mobile applications.
\newblock In {\em Proceedings of the 2016 ACM on Multimedia Conference}, 2016.

\bibitem{Wen_NIPS2016}
W.~Wen, C.~Wu, Y.~Wang, Y.~Chen, and H.~Li.
\newblock Learning structured sparsity in deep neural networks.
\newblock In {\em Advances in Neural Information Processing Systems (NIPS)}.
  2016.

\bibitem{zeiler2014visualizing}
M.~D. Zeiler and R.~Fergus.
\newblock Visualizing and understanding convolutional networks.
\newblock In {\em European Conference on Computer Vision (ECCV)}, 2014.

\bibitem{7332968}
X.~Zhang, J.~Zou, K.~He, and J.~Sun.
\newblock Accelerating very deep convolutional networks for classification and
  detection.
\newblock {\em IEEE Transactions on Pattern Analysis and Machine Intelligence},
  38(10):1943--1955, Oct 2016.

\end{thebibliography}
}

\end{document}